\documentclass[runningheads]{llncs}

\usepackage[mobile]{eccv}

\usepackage{eccvabbrv}

\usepackage{graphicx}
\usepackage{booktabs}

\usepackage[accsupp]{axessibility}  

\usepackage[misc]{ifsym}

\usepackage{hyperref}

\usepackage{orcidlink}
\usepackage[dvipsnames]{xcolor}

\usepackage[titletoc,toc,page]{appendix}

\definecolor{cvprblue}{rgb}{0.21,0.49,0.74}
\usepackage{soul}
\usepackage{multirow}

\usepackage{tabularx}

\newcolumntype{L}{>{\raggedright\arraybackslash}X}
\newcolumntype{C}{>{\centering\arraybackslash}X}
\newcolumntype{R}{>{\raggedleft\arraybackslash}X}

\begin{document}

\title{SphereHead: Stable 3D Full-head Synthesis with Spherical Tri-plane Representation} 


\titlerunning{SphereHead: Stable 3D Full-head Synthesis}

\author{Heyuan Li\inst{1}\orcidlink{0000-0001-8101-2916} \and
Ce Chen\inst{1}\orcidlink{0009-0001-8267-8346} \and
Tianhao Shi\inst{1}\orcidlink{0009-0002-6795-1452} \and
Yuda Qiu\inst{1}\orcidlink{0009-0003-1257-4271} \and
Sizhe An\inst{3}\orcidlink{0000-0002-9211-4886} \and
\\
Guanying Chen\inst{1}\orcidlink{0000-0002-1273-4752} \and
Xiaoguang Han\inst{1,2}\href{mailto:hanxiaoguang@cuhk.edu.cn}{\textsuperscript{\Letter}}\orcidlink{0000-0003-0162-3296}
}


\authorrunning{H.~Li, C.~Chen, T.~Shi, Y.~Qiu, S.~An, G.~Chen, and X.~Han.}

\institute{
School of Science and Engineering, The Chinese University of Hong Kong, Shenzhen
\and The Future Network of Intelligence Institute, CUHK-Shenzhen
\and University of Wisconsin-Madison
}

\maketitle
\begin{center}
\centering
    \vspace{-2mm}
    {\includegraphics[width=1.0\linewidth]{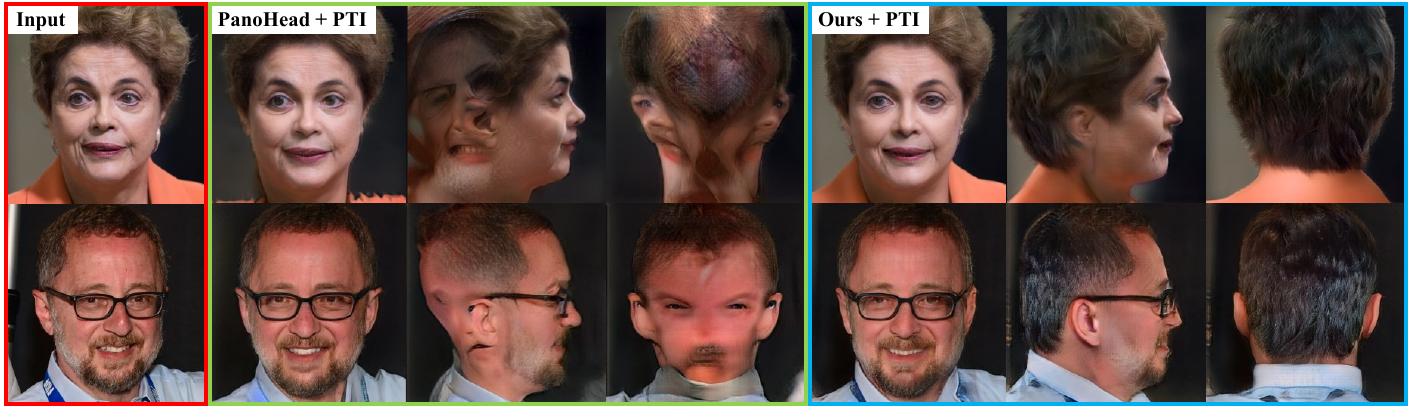}}
    \vspace{-6mm}
    \captionof{figure}
    {
    Given input images, applying PTI\cite{roich2022pivotal} methods to reconstruct full-head synthesis based on PanoHead\cite{an2023panohead} often yields noticeable artifacts (green box), while those using our SphereHead space (blue box) appear more realistic.
    }
    \label{fig:teaser}
\end{center}
\vspace{-4mm}

\begin{abstract}

While recent advances in 3D-aware Generative Adversarial Networks (GANs) have aided the development of near-frontal view human face synthesis, the challenge of comprehensively synthesizing a full 3D head viewable from all angles still persists. Although PanoHead \cite{an2023panohead} proves the possibilities of using a large-scale dataset with images of both frontal and back views for full-head synthesis, it often causes artifacts for back views. Based on our in-depth analysis, we found the reasons are mainly twofold. 
First, from network architecture perspective, we found each plane in the utilized tri-plane/tri-grid representation space tends to confuse the features from both sides, causing ``mirroring'' artifacts (e.g., the glasses appear in the back).   
Second, from data supervision aspect, we found that existing discriminator training in 3D-aware GANs mainly focuses on the quality of the rendered image itself, and does not care much about its plausibility with the perspective from which it was rendered. This makes it possible to generate ``face'' in non-frontal views, due to its easiness to fool the discriminator.
In response, we propose SphereHead, a novel tri-plane representation in the spherical coordinate system that fits the human head's geometric characteristics and efficiently mitigates many of the generated artifacts. We further introduce a view-image consistency loss for the discriminator to emphasize the correspondence of the camera parameters and the images. 
The combination of these efforts results in visually superior outcomes with significantly fewer artifacts. 
Our code and dataset is publicly available at \url{https://lhyfst.github.io/spherehead/}.

\vspace{-2mm}

\keywords{Head Synthesis \and 3D GANs \and Spherical Tri-plane}

\end{abstract}

\section{Introduction}
\label{sec:intro_new}

The synthesis of human head portrait images has long been of great concern within the realms of computer vision and graphics, with extensive applications spanning video conferencing, digital avatars, games, movies, and more. While 2D GANs are well-studied and can generate photorealistic facial images, they encounter challenges when applied to multi-view scenarios, where the relative position and orientation of the camera and the head are in flux. 
The recent developments in neural implicit representation have paved the way for the genesis of 3D-aware GANs, which achieve photorealistic view-consistent image synthesis. A groundbreaking work in this realm is EG3D \cite{chan2022efficient}, which puts forth an efficient and effective tri-plane representation in Cartesian space. This approach has been widely adopted by numerous 3D-aware synthetic head image generators that followed. 
However, these generators are restricted to near-frontal synthesis and are unable to produce back-view synthesis due to the scarcity of high-quality multi-view data.

\begin{figure}[t]
  \centering
   \includegraphics[width=1.0\linewidth]{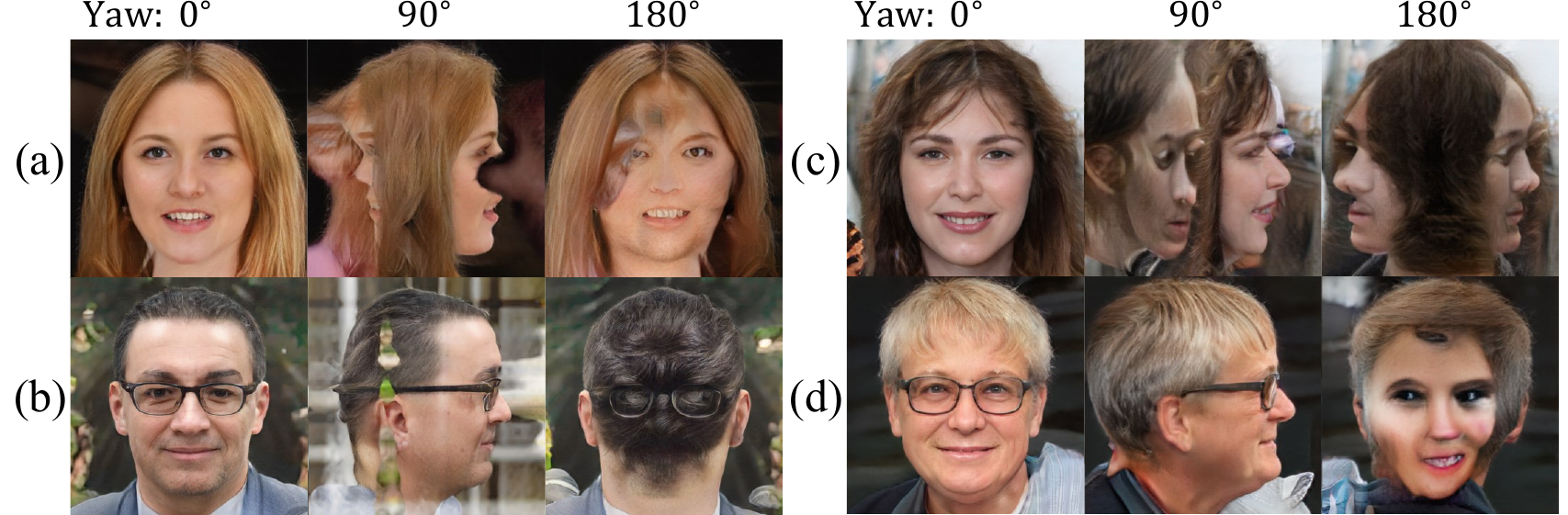}
       \caption
       {
       Two types of fake face artifacts in PanoHead. (a-b) We name the first type as \textit{mirroring-face artifacts}, due to the back face mirroring the identity, expression and accessories of the front face precisely.
       (c-d) We name the second type as \textit{multiple-face artifacts}, because in this scenario there might be more than one fake faces and their identities, expression and accessories are different from the front face. 
       }
   \label{fig:intro}
\end{figure}

A comprehensive 3D head synthesis, viewable from all angles, is pivotal for enhancing immersive experiences, particularly in interactive scenarios such as 3D video games and digital avatars. PanoHead \cite{an2023panohead}, the first 3D-aware GAN for full head synthesis, addresses this need to an extent. It enables the training with monocular back-view images using an adaptive camera strategy and extends the tri-plane representation into a tri-grid version for back-view synthesis. 
While the results affirm the feasibility of full-view synthesis without 3D data, they frequently exhibit noticeable semantic mismatch artifacts, with the most pronounced being the fake face on the back-view head, as depicted in ~\cref{fig:intro}.
This can be traced back to two aspects. 

From the network architecture perspective, representations based on the Cartesian coordinate system, such as tri-plane and tri-grid, naturally introduce feature entanglement for symmetrical areas relative to the predefined planes.
For example, the back-view head shares the same features on `XY' plane with the front-view face, resulting in \textit{mirroring-face artifacts} on the back of head or unreasonable neck shape, as depicted in ~\cref{fig:intro} (a, b).

From the data supervision aspect, in 3D-aware generation, it's crucial to ensure the correspondence between the generated images and their viewing perspectives. 
However, while existing 3D-aware GANs take both images and camera parameters as inputs for the discriminator, they fail to clearly distinguish between two unwanted cases: unrealistic images paired with their appropriate camera parameters, and realistic images accompanied by incorrect camera parameters.
This causes the discriminator tends to score the generated data pairs by the visual quality of the images and weakens the consistency between the images and their viewing perspectives.
This becomes especially problematic in scenarios with marked imbalances in supervision intensity across different directions, such as with datasets of head images. 
As a result, the generator tends to create heavily supervised content, like faces, in incorrect orientations, resulting in \textit{multiple-face artifacts} as depicted in ~\cref{fig:intro} (c, d).

In this paper, we propose a novel framework, SphereHead, to enhance panoramic head image synthesis. Our method introduces an entanglement-free representation that reduces mirroring artifacts. In particular, regarding the geometric characteristics of the human head, we propose a dual spherical tri-plane representation which is defined in the spherical coordinate system. This sphere's geometric properties explicitly separate the features of each position on the head, thereby resolving mirroring artifacts completely. 

Moreover, we introduce a view-image consistency loss to enforce the discriminator focusing on identifying unaligned pairs of images and camera parameters, even if the quality of images are exceedingly high. We achieve this by intentionally pairing real images with shuffled real camera parameters, and designating these pairs as negative samples for the training of discriminator. With such additional supervision from negative real pairs, the discriminator manages to judge an image not only by its fidelity but also by the plausibility of its viewing perspective. 
Thereby, we successfully tackle the two types of semantic mismatch artifacts by the spherical tri-plane representation and the view-image consistency loss respectively.

Additionally, although an in-house dataset of back-view head images plays a crucial role in PanoHead, it remains inaccessible to the research community. 
In response, we have compiled a large-scale dataset featuring 60k head images from various viewpoints, including the back view. This dataset was developed through extensive data collection and thorough cleaning processes. It not only supports our research but is also publicly available, along with a comprehensive data processing toolbox.


Through extensive qualitative and quantitative experiments, our method demonstrates an improved ability to produce 3D full head synthesis. These results exhibit enhanced visual quality and significantly fewer artifacts compared to existing methods. In summary, the main contributions of this paper are:

\begin{itemize}
    \item We proposed a novel framework, SphereHead, which significantly reduces the noticeable artifacts of PanoHead, and effectively addresses the challenge of lacking 3D data for stable full-head synthesis. 
    \item We developed a novel spherical tri-plane representation, which effectively solved the feature entanglement issue of existing regular tri-plane/tri-grid representation. 
    \item We introduce a view-image consistency loss to compel the discriminator to focus on the alignment between images and their viewpoints, thereby minimizing artifacts arising from imbalanced supervision.
    \item We collected a large-scale head image dataset, the WildHead Dataset, whose images are captured in the wild from various viewing angles spanning 360 degrees. This dataset is publicly available on our project page.

\end{itemize}

\section{Related Work}
\label{sec:related}

\noindent \textbf{3D Head Modeling.} 
Considerable research has focused on explicitly representing a 3D human head using textured mesh, such as 3DMM \cite{blanz1999morphable}, BFM \cite{paysan20093d} and FLAME \cite{li2017learning}. 
These models typically ascertain the distribution of shape and texture variations from a substantial collection of 3D head scans, employing Principle Component Analysis (PCA).
Nonetheless, their pre-defined topologies limit the capture of details and accessories such as hair, glasses, and wrinkles, leading to the generation of unrealistically rendered images. 
In the other line, an increasing number of recent 3D head synthesis methods have adopted neural implicit representation due to its high expressivity. Many such studies \cite{zheng2022imface, canela2023instantavatar, zanfir2022phomoh} represent 3D heads via Signed Distance Functions (SDFs), which model 3D shapes by defining the signed distance from every 3D location to its nearest surface. Another group of works \cite{hong2022headnerf, yenamandra2021i3dmm, wu2023ganhead, zhuang2022mofanerf, gafni2021dynamic, guo2021ad, park2021nerfies, zhang2023metahead} represents the 3D head using Neural Radiance Fields (NeRFs) \cite{mildenhall2021nerf}, which map a 3D location and viewing direction to a color and a density value, then synthesize images through volume rendering.

\noindent \textbf{Generative 3D-Aware Image Synthesis.} 
3D-aware GANs, learning from collections of 2D images, aim to synthesize view-consistent images from various viewpoints by capturing the inherent 3D geometry and appearance of the scene.
While earlier literature explicitly encodes the 3D scene using mesh \cite{szabo2019unsupervised, shi2021lifting, liao2020towards} and voxel grids \cite{gadelha20173d, nguyen2020blockgan, nguyen2019hologan}, a growing number of studies \cite{deng2022gram, or2022stylesdf, chan2021pi, schwarz2020graf} utilize implicit representations which are more expressive, thereby demonstrating high-quality image generation capabilities. 
Recently, EG3D \cite{chan2022efficient} puts forward a tri-plane implicit-explicit hybrid 3D representation, which later works have extensively adopted, notably for its efficiency. 
PanoHead \cite{an2023panohead} enriched the tri-plane's representational capacity through adding more parallel feature planes, thereby lessening the artifacts in back-view head image synthesis. Nonetheless, persistent artifacts endure, as its tri-grid representation retains a similar geometric structure, which is fundamentally the source of the artifacts.
In contrast, our method, through a shift in formulation from a Cartesian coordinate representation in cubic space to a spherical coordinate representation in spherical space, avoids many artifacts at their root.

\noindent \textbf{View-Image consistency in 3D-aware GAN} Inspired by cGAN\cite{mirza2014conditional}, \cite{gauthier2014conditional,denton2015deep} explore the training of conditional GAN for image generation. They directly add the conditional labels for both of the generator G and the discriminator D. The EG3D-like methods work in the same manner to maintain the view-image consistency. However, the experiments show the networks fail to ensure the consistency while training in dataset with biased distribution. To further compel D to identify image-label matching in addition to the image realism, \cite{reed2016generative} propose to train D with additional negative real pairs consisting of real images with mismatched labels. We adapt this as a view-image consistency loss to emphasize the correspondence of the camera labels and the images in our 3D-aware synthesis framework.

\noindent \textbf{Human Head Portrait Datasets.} 
To ensure the diversity and quality of synthetic images, generative models necessitate learning from large-scale, high-quality portrait images. CelebA \cite{liu2015faceattributes} and FFHQ \cite{karras2018style} are two of the most widely-used single-view portrait datasets, encompassing 200,000 and 70,000 images respectively. However, as the majority of their images are captured from narrow, near-frontal views, they are inadequate for training a full head generator.
The LPFF dataset \cite{Wu_2023_ICCV}, while offering a larger viewing angle for portrait images, is still deficient in back-view images. The distribution bias in these datasets impedes the learning of 3D full head synthesis. Following the practice in PanoHead\cite{an2023panohead}, we collect a dataset with back-view head images to enable the  training of full head generative model.

\section{SphereHead}
\label{sec:method}

Our novel SphereHead extends upon the full-head synthesis work of PanoHead by incorporating a dual spherical tri-plane representation (Sec.~\ref{sec:method_rep}), a view-image consistency loss (Sec.~\ref{sec:method_neg}), and a parsing branch (Sec.~\ref{sec:method_par}). 

\begin{figure}
\centering 
    {\includegraphics[width=1.0\linewidth]{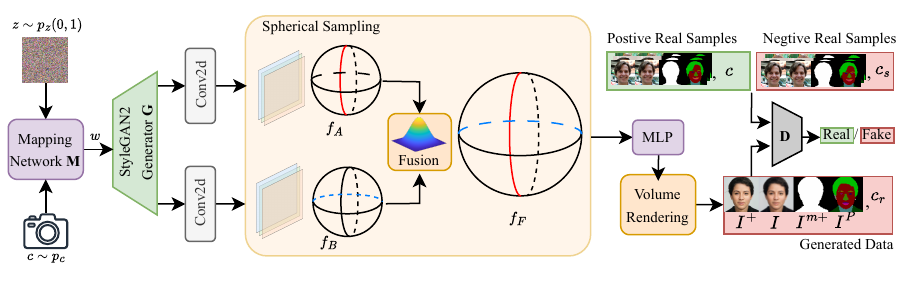}}

    \caption{The framework of our proposed SphereHead. Given a sampled code $z$ and camera parameter $c$, SphereHead synthesizes a spherical tri-plane features $f_F$ by fusing two sub-feature groups $f_A$ and $f_B$. By volumetric rendering with the sampled features in $f_F$, SphereHead generates high-quality view-consistent full head images $I^{+}$. The discriminator learns to focus on the alignment between images and their viewpoints instructed by our view-image consistency loss, by introducing an additional negative data pairs consisting the real images and mismatched labels $c_s$.}
    \label{fig:main_pipeline}
\end{figure}

As depicted in Fig.~\ref{fig:main_pipeline},
our generator $G$ comprises a modified StyleGAN2 generator, augmented with two additional sub-modules. These modules synthesize two distinct feature groups $f_A$ and $f_B$, each consists of three 2D feature planes: $P_{\theta r}$, $P_{\phi r}$, and $P_{\theta\phi}$ corresponding to the spherical coordinate system.
We merge $f_A$ and $f_B$ into a fused tri-plane feature $f_F$ using a carefully designed weight map. 
With $f_F$ and the rendering camera poses $c_{r}$, a neural renderer $R$ generates a set of low-resolution attribute maps. These maps are further processed into a super-resolved image $I^{+}$, a bilinear upsampled image $I$, a super-resolved foreground mask $I^{m+}$, and a bilinear-resolved face parsing $I^{p}$. 
The discriminator $D$ then critiques generated $(I^{+}, I, I^{m+}, I^{p})$ and $c_{r}$, alongside real images paired with matching and mismatched real camera parameters.

\subsection{Dual Spherical Tri-plane Representation}
\label{sec:method-dual}

The primary challenge in 3D full-head synthesis lies in generating a plausible occipital region, particularly when dealing with complex and diverse hairstyles. On the one hand, the fundamental issue stems from biased supervision arising from imbalanced training data. Specifically, the abundance of high-quality frontal view data fosters strong supervision for frontal view synthesis learning. In contrast, back view synthesis learning is hindered due to deficient back view data, often marred by unreliable camera pose estimation, insufficient high-quality images, and subpar image alignment. This data imbalance impedes the learning process for back view synthesis, resulting in numerous artifacts on the occipital region.

On the other hand, the prevalent tri-plane representation is intrinsically prone to fostering feature entanglement due to its inherent structure and Cartesian feature projection. 
Specifically, the same feature on a plane is symmetrically shared by disparate positions about the plane via Cartesian projection, leading to undesirable mirroring artifacts. This issue becomes especially pronounced within the occipital region due to the imbalanced supervision between the frontal and back views. The facial features from the front view, which have been powerfully supervised, dominate the $P_{XY}$ feature map and permeate into the poorly supervised occipital region, leading to striking mirroring-face artifacts. 
Although PanoHead's tri-grid representation mitigates this issue to a degree by incorporating additional feature planes, it retains a similar structure and operates within the same Cartesian coordinate system, leading to the persistence of mirroring artifacts in practice.

\label{sec:method_rep}
\begin{figure}[h!]
    \centering
    \includegraphics[width=0.9\linewidth]{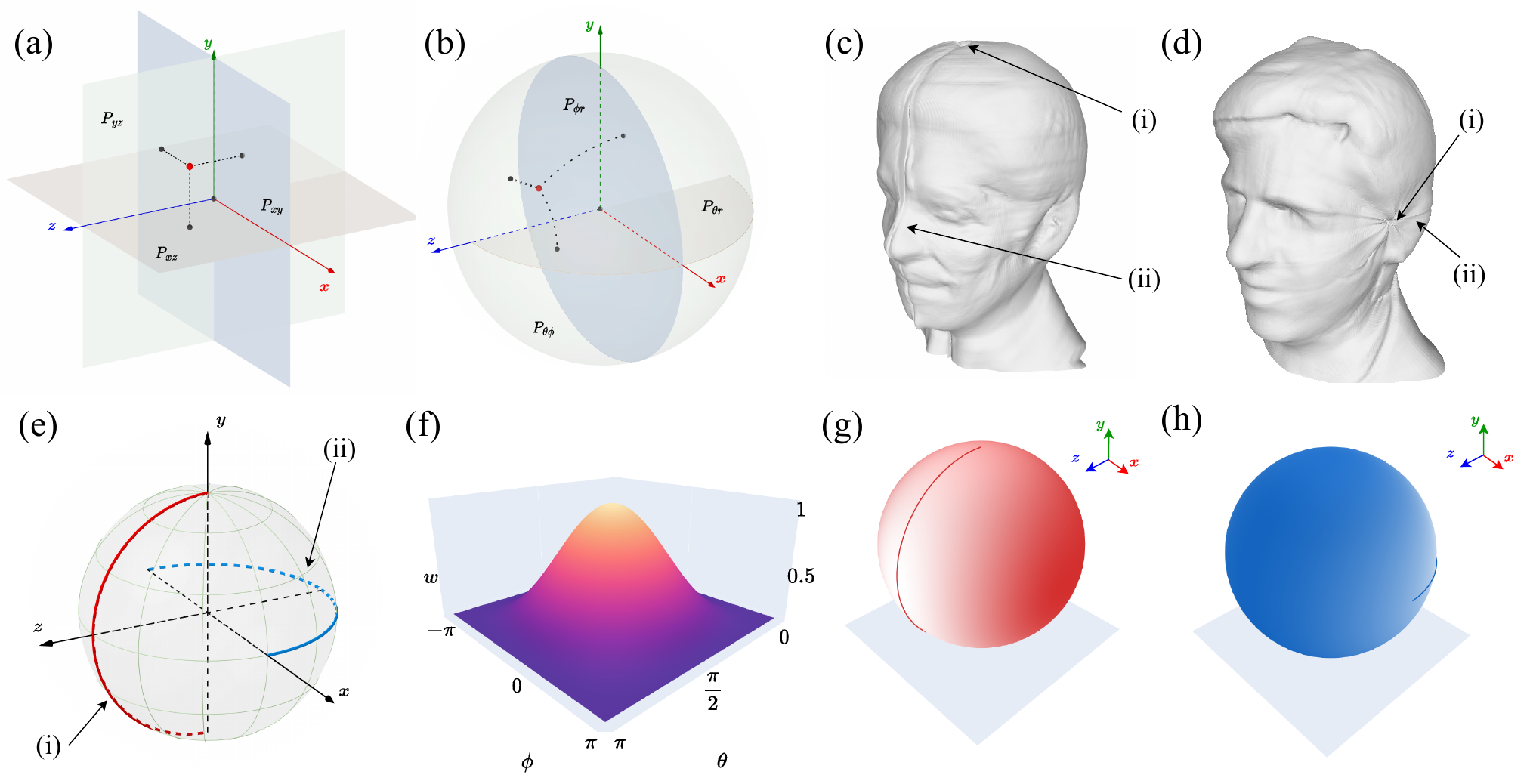}
    \caption{(a) Tri-plane representation. (b) Spherical tri-plane representation. Reconstructed head geometry from single (c) sphere A and (d) sphere B, each showing (i) seam artifacts and (ii) polar artifacts. (e) The combination of two spheres in dual spherical tri-plane representation, (i) the seam of sphere A, (ii) the seam of sphere B. (f) Fusion weight map. (g-h) For each sphere, the weight approaches zero as the locations near the seam and poles.
    }
    \label{fig:method_dual}
\end{figure}

To this end, we propose a novel spherical tri-plane representation, which explicitly separates features from different facial regions. 
Specifically, we represent a point in the spherical coordinate system as $(r, \theta, \phi)$.
Correspondingly, as illustrated in Fig.~\ref{fig:method_dual}~(b), the three feature planes in our spherical tri-plane representation embody a semicircle plane $P_{\theta r}$, a circular plane $P_{\phi r}$, and a spherical plane $P_{\theta \phi}$. 
We query a 3D position $x \in \mathbb{R}^{3}$ by projecting it onto each of the three feature planes via its spherical coordinate, retrieving the respective feature vector $(F_{\theta r}, F_{\phi r}, F_{\theta \phi})$ through bilinear interpolation, and subsequently merging them via summation. 
A lightweight decoder network interprets the combined 3D features $F$ as color and density, which are used to render RGB images through volume rendering.

Our spherical tri-plane representation is demonstrated to entirely eliminate mirroring artifacts referring to Sec.~\ref{sec:expr_qual}. 
However, in practice, these three feature planes are wrapped from square feature maps, which leads to potential issues in $P_{\theta \phi}$ and subsequently, the emergence of new artifacts in the seam and polar regions, clearly depicted in Fig.~\ref{fig:method_dual}~(c, d). 
The former stems from the numerical discontinuity of $F_{\theta \phi}$ at $\phi=-\pi$ and $\phi=\pi$. Meanwhile, the latter comes to light when $\theta$ nears 0 and $\pi$, leading to the convergence of $\phi=-\pi$ and $\phi=\pi$ lines to two polar points. 
Such transformation converts the numerical fluctuations on these lines into high-frequency noise around the poles. 

To contend with these complications, we introduce an additional spherical tri-plane, orthogonal to the existing one, i.e. their polar axes are mutually perpendicular, and seam openings are oriented in opposite directions, which is shown in Fig.~\ref{fig:method_dual}~(e). This complementarity is encapsulated by a 2D weighting map, given by:
\begin{equation}
w(\theta,\phi)=\frac{1+\cos (2\theta-\pi)}{2} \cdot \frac{1+\cos (\phi)}{2}.
\label{equ:weight}
\end{equation}
As indicated in Fig.~\ref{fig:method_dual}~(f-h), $w$ attains maximum value at its center and gradually recedes to 0 at its edges, where the seam and poles are. Due to this arrangement, the two planes wrap around each other akin to the two leathers of a baseball, each utilizing its smooth region to compensate for the problematic areas of the other. By implementing this strategy, we successfully eradicate the seam and polar artifacts.

\subsection{View-image Consistency Loss (ViCo Loss)}

\label{sec:method_neg}

In addition to mirroring artifacts, the imbalanced supervision also gives rise to the multiple-face artifacts depicted in \cref{fig:intro}~(c, d).
In these instances, the fake back faces exhibit divergent attributes compared to the frontal faces, indicating that they are not the result of feature entanglement.

This supervisory imbalance skews the generator's ability to synthesize high-quality frontal faces, while significantly impairing its capacity to generate satisfactory posterior views.
As detailed in \cite{denton2015deep}, in early training, discriminators tend to assess authenticity based on image quality rather than fidelity to specific attributes.
Consequently, to deceive the discriminator, the generator opts to replicate facial features even in non-frontal images, as this approach more readily enhances visual quality than the more challenging task of accurately generating the occipital region. 
This compulsion to enhance visual quality by facial features undermines the authentic generation of the posterior aspect of the head, resulting in a consistent production of artifact-laden back-view images that fall short of real image realism.

This loop creates a negative feedback mechanism: the persistent artifacts prevent the discriminator from evolving its criteria from mere visual quality to a more nuanced analysis of the congruence between images and their corresponding viewpoints. Consequently, the training process becomes ensnared in a dilemma, undermining the model's overall ability to generate convincing images across varying viewpoints.

Our view-image consistency loss is designed to disrupt this negative feedback loop by compelling the discriminator to simultaneously evaluate input images based on both their view-image consistency and visual quality. 
In practice, we construct negative pairs using real data to facilitate our view-image consistency loss. 
Specifically, for a batch of input data $\{(I^{+}, I, I^{m+}, I^{p}),c\}$, we shuffle the camera labels $c$ to obtain $c_s$. The mismatched negative pairs $\{(I^{+}, I, I^{m+}, I^{p}), c_s\}$ are then fed into the discriminator $D$, as illustrated in ~\cref{fig:main_pipeline}. 
The view-image consistency loss for the discriminator is formulated as following:
\begin{equation}
    \mathbb{L}_{ViCo} = log(1-D((I^{+}, I, I^{m+}, I^{p}),c_s)).
\end{equation}
Thus, the discriminator can separate supervision by direction, thereby protecting poorly supervised directions from the undue influence of those with stronger supervision.
As a result, the generator benefits from purer back-view supervision and steadily enhancing the visual and content accuracy of the back-view image synthesis.

\subsection{Parsing Branch for Semantic Embedding}
\label{sec:method_par}

In the renderer $R$, along with color and density, we additionally obtain parsing feature from 3D features $F$, which is further rendered into face parsing map $I^{p}$. 
Although previous works \cite{zhou2023lc, sun2022fenerf, jiang2022nerffaceediting, zhang2023metahead} exploit a similar parsing branch for controlled synthesis, they do not consider its potential contribution to generator learning. However, our experiments (Sec.~\ref{sec:expr_quan}) indicate that learning facial parsing benefits the generator in terms of visual quality and artifacts reduction. The learning of low-frequency information in face parsing enhances the semantic understanding of each face region, driven the generator to smooth the features in the same region while distinguish those from different regions.

\section{Dataset}
\label{sec:dataset}

\noindent \textbf{WildHead Dataset.}
Although PanoHead has crafted a dataset of back-view head images, its exclusivity has presented a challenge for the broader research community. 
To address this issue, we present the first publicly accessible large-scale dataset, the WildHead Dataset, whose images are captured in the wild from various viewing angles spanning 360 degrees.

\noindent \textbf{A Collection of Head Images for Full-Head Synthesis.} 
We incorporate our dataset with publicly available datasets, including CelebA \cite{liu2015faceattributes}, FFHQ \cite{karras2018style}, and LPFF \cite{Wu_2023_ICCV}, to train our full-head synthesis models. 
Besides, we also utilize partial back-view images in K-Hairstyle \cite{kim2021k} dataset to enhance the back-view hairstyle diversity of the training set. 
To improve data quality, we carefully design a pipeline to filter and process all the images from the gathered dataset. 
Additionally, meticulous manual vetting has been employed to ensure the high quality of the dataset.
As a result of these rigorous curation efforts, the final dataset comprises a total of 180k images.

\noindent \textbf{Data Processing and Open-Source Toolbox.}
In order to train our model with the collected data, it is necessary to align the images and obtain the corresponding camera parameters. 
Following PanoHead~\cite{an2023panohead}, we align the portraits to the center of images, with the aid of facial landmarks detector~\cite{bulat2017far} for near-frontal face images and Yolo Head Detector \cite{zhang2021soft} for others. Then we estimate the extrinsic camera parameters with the assumption that all cameras have the same intrinsic matrix and are facing towards the origin, resulting in the camera labels of our dataset. 
To mitigate the imbalance of perspectives within the dataset, we replicated the images from sparsely represented viewpoints multiple times.
The entire data processing pipeline has been packaged as a toolbox which is publicly available in our released code. Please refer to the supplementary materials for more detailed information.

\section{Experiments}
\label{sec:exper}

\begin{figure}
\centering
    {\includegraphics[width=\linewidth]{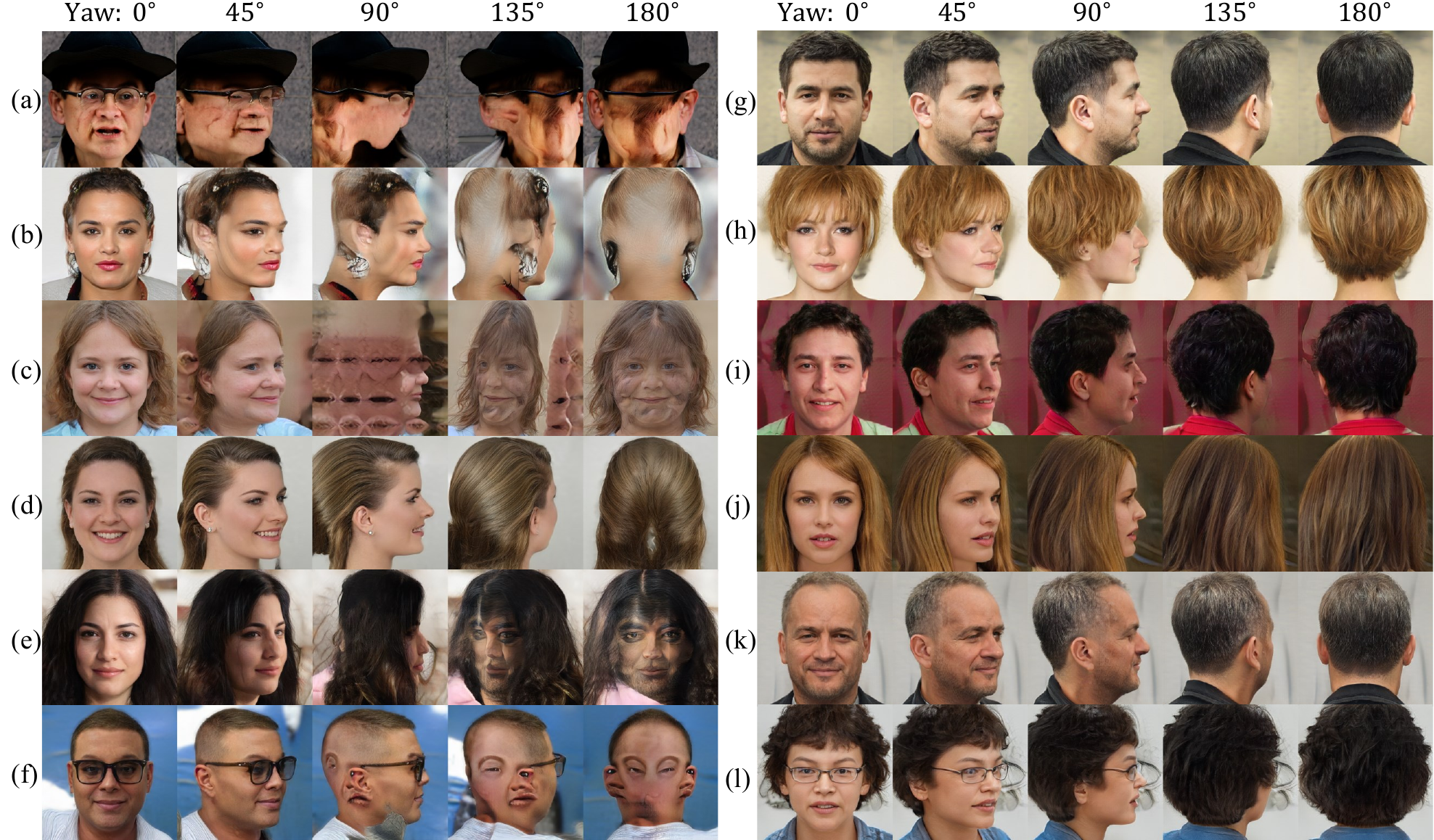}}
    \vspace{-6mm}
    \caption{Qualitative comparison with state-of-the-art methods. (a) GIRAFFHD \cite{xue2022giraffe}, (b) StyleSDF \cite{or2022stylesdf}, (c) EG3D \cite{chan2022efficient} fail to capture the complete head geometry and appearance. (d-f) PanoHead \cite{an2023panohead} shows complete head generation, but the results suffer from mirroring artifacts ((d) left-right identical mirroring artifacts and (f) mirroring-face artifacts) and (e) multiple-face artifacts). (g-l) Ours SphereHead synthesizes full-head images of high visual quality and is free of artifacts exhibited by other methods.}
    \label{fig:expr_360}
    \vspace{-4mm}
\end{figure}

\subsection{Training Details and Baselines}

Our model is trained on eight NVIDIA A100 GPUs with a batch size of 32. 
The resolution for both the images within the training set and the subsequently synthesized images measures $512^2$.
Over the span of 7 days, the thorough training process encompasses three distinct phases with 25 million images.

\noindent \textbf{Phase I} (0-2M images): 
our model is exclusively trained with near-front view images for their superior quality, which facilitates a robust initialization. 
We adopt an alternative training strategy, which randomly trains $f_A$, $f_B$, and $f_F$ with the probabilities of 33\%, 33\%, and 34\% respectively.
This approach accelerates the learning of both spherical tri-planes and curtails the formation of evident artifacts. 
The view-image consistency loss is not applied in the beginning, as it may hinder early generator training to improve image quality.
We blur images as they enter the discriminator, gradually reducing the blur amount over this duration.

\noindent \textbf{Phase II} (2-10M images): We begin to train our model on the entire training dataset.

\noindent \textbf{Phase III} (10M-25M images): We integrate the view-image consistency loss in this stage. 
By re-calibrating the training probabilities for $f_A$, $f_B$, and $f_F$ to 10\%, 10\%, and 80\% respectively, we enforce the generator to focus on the output of the fused sphere.

We compare against state-of-the-art 3D-aware GANs including GIRAFFEHD \cite{xue2022giraffe}, StyleSDF \cite{or2022stylesdf}, EG3D \cite{chan2022efficient}, and PanoHead \cite{an2023panohead}. Unless otherwise indicated, all baselines are trained on our comprehensive head image collection. We evaluate the quality of the generated multi-view images, quantitatively and qualitatively.

\subsection{Qualitative Comparisons}
\label{sec:expr_qual}

\noindent \textbf{360$^{\circ}$ Image Synthesis.} 
Fig.~\ref{fig:expr_360} offers a visual comparison of image synthesis among our method and the baselines, presenting synthesized images from five distinct views with the yaw angle ranging from 0 to 180$^{\circ}$. 
GIRAFFEHD and StyleSDF manage to produce photo-realistic front-view synthesis, but failed to generate reasonable images on larger viewing angles since their rendering does not explicitly consider camera poses. 
EG3D can render back-views to some extent. However, it tends to confuse the features of the frontal view with those of the back-view region, leading to mirroring face artifacts. Additionally, it suffers from floating artifacts due to its lack of proper foreground and background separation.
PanoHead, as the first work targeting 3D full-head synthesis, delivers superior 360$^{\circ}$ synthesis results compared to other baseline methods. Nevertheless, PanoHead still struggles with outputs with mirroring features and multiple-face artifacts. 
In contrast, our proposed solution, SphereHead, consistently produces high-quality, artifact-free 360$^{\circ}$ synthesis, demonstrating its superiority over pre-existing methods.

\begin{figure}[t]
    \centering
    \includegraphics[width=\linewidth]{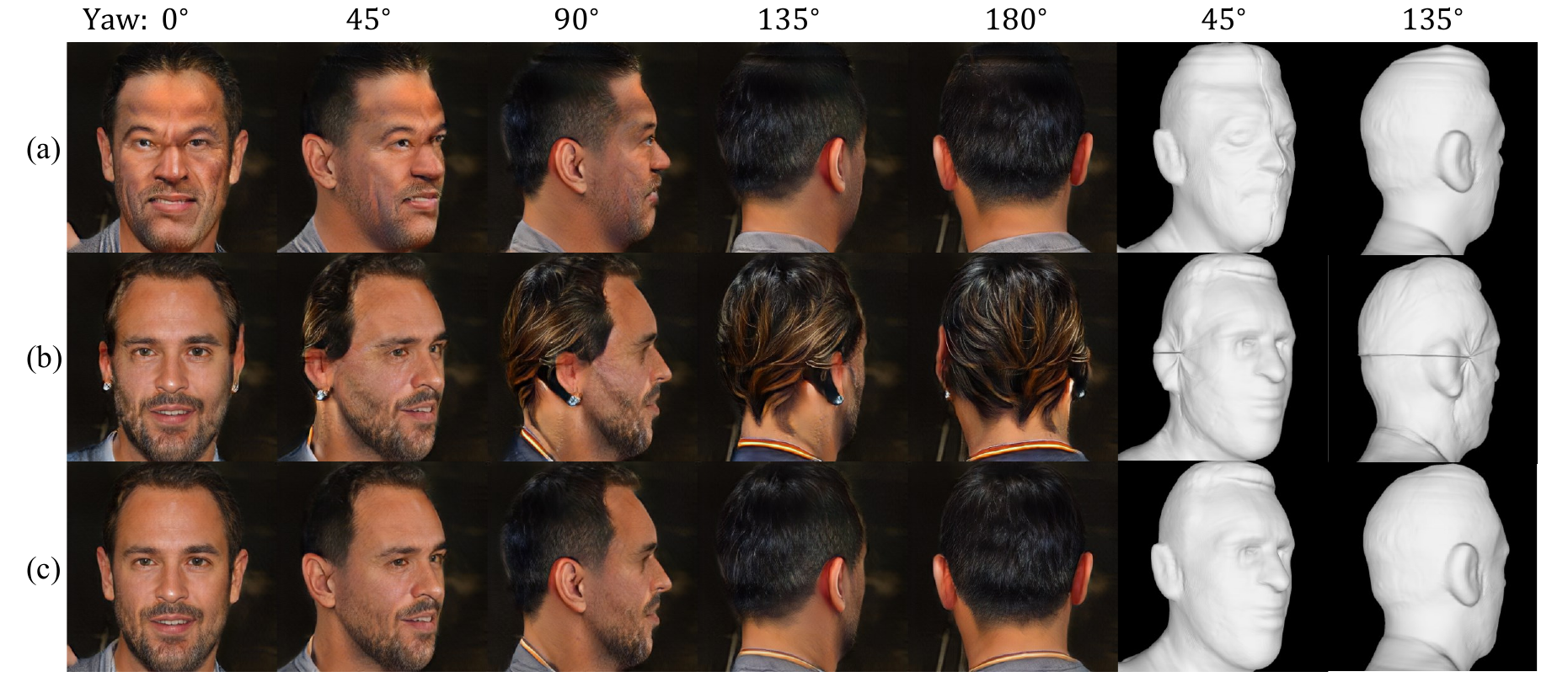}
    \caption{Synthetic images and geometries rendered by (a) only $f_A$, (b) only $f_B$ and (c) fused $f_F$. While the individual spherical space exhibits notable seam and polar artifacts, the fused $f_F$ successfully mitigates the artifacts. }
    \label{fig:exper_dual}
\end{figure}

\noindent \textbf{Dual Sphere Fusion.} 
We evaluate our dual spherical tri-plane design by rendering synthesized images and corresponding geometries separately from features $f_A$, $f_B$, and their fused $f_F$, as demonstrated in Fig.~\ref{fig:exper_dual}. The results indicate that the synthesized image and geometry of frontal view are dominated by $f_A$, while those of back-view region are primarily influenced by $f_B$.
Employing our strategically developed weight map, these two features effectively integrate and support each other. While each individual spherical features exhibits notable seam and polar artifacts, $f_F$ strategically masks the flaws by fusing the artifact-free areas in $f_A$ and $f_B$. Consequently, our design successfully mitigates individual limitations, leading to an enhanced final result.

\subsection{User Study}
To fairly and quantitatively assess the capability of our method in reducing semantic mismatch artifacts, we conducted a rigorous user study. 
Within this study, we established two distinct metrics. A collection of 100 random samples from each model's latent space served as the test subjects. 37 participants were tasked with making two determinations for each sample: firstly, whether the back-view images exhibited any discernible fake facial artifacts, and secondly, whether the sample appeared artifact-free from all angles, indicating a high-quality result. The frequency of identifiable fake face artifacts across all samples was denoted as the \textit{fake face rate (FF)}, while the prevalence of samples deemed artifact-free was referred to as the \textit{high-quality rate (HQ)}. The resultant data from these evaluations contributed to both the qualitative comparisons in \cref{tab:exper_baseline} and the ablation studies in \cref{tab:exper_ablation}, offering insights into our method's effectiveness in artifact reduction.

\subsection{Quantitative Results}
\label{sec:expr_quan}

\paragraph{Comparisons with SOTA} In line with prior researches on 3D-aware GANs, we employ the Frechet Inception Distance (FID) calculated from 50,000 real and synthesized image samples to quantitatively evaluate the visual quality, fidelity, and diversity of the generated images. 
Evidenced by possessing the lowest FID and FF, as well as the highest HQ as shown in \cref{tab:exper_baseline}, our SphereHead method surpasses other state-of-the-art methods in terms of image quality and artifact reduction.

\begin{table}[th]
    \centering
    \setlength{\tabcolsep}{3mm}
        \begin{tabular}{llllll}
        \toprule
            & \cite{xue2022giraffe} & \cite{or2022stylesdf} & EG3D & PanoHead  & Ours \\ \cline{2-6}
            FID $\downarrow$ & 67.8 & 105.5 & 10.3  & 8.6  & \textbf{7.8} \\
            FF $\downarrow$ & \textendash{} & \textendash{} & 76.1\%  & 23.7\% & \textbf{0\%} \\
            HQ $\uparrow$ & \textendash{} & \textendash{} & 3.7\% &26.1\% & \textbf{92.6\%} \\
        \bottomrule
        \end{tabular}
    \caption{Quantitative comparison among our method and state-of-the-art 3D-aware GANs. 
    GIRAFFEHD\cite{xue2022giraffe} and StyleSDF\cite{or2022stylesdf} are excluded from the user study because they completely fail to generate reasonable back-view head images.}
    \label{tab:exper_baseline}
\end{table}
\vspace{-3mm}

\begin{table}[th]
    \centering
    \small
    \begin{tabular}{p{2cm}<{\centering}p{2cm}<{\centering}p{2cm}<{\centering}|p{1.5cm}<{\centering}p{1.5cm}<{\centering}p{1.5cm}<{\centering}}
    \toprule
        Representation & ViCo Loss  & Parsing & FID $\downarrow$ & FF $\downarrow$ & HQ $\uparrow$\\ \midrule
        Tri-grid & \textendash{} & \textendash{} & 8.6 & 23.7\% & 26.1\% \\
        Tri-grid & \checkmark & \textendash{} & 8.3 & 4.8\% & 63.2\% \\
        Single-Sphere & \textendash{} & \textendash{} & 8.1 & 3.7\% & 78.7\%\\
        Single-Sphere & \checkmark & \textendash{} & 7.9 & 0.3\% & 85.6\%\\
        Dual-Sphere & \textendash{} & \textendash{} & 7.9 & 0.9\% & 84.2\%\\
        Dual-Sphere & \checkmark & \textendash{} & 7.9 & \textbf{0\%} & 90.4\%\\
        Dual-Sphere & \checkmark & \checkmark & \textbf{7.8} & \textbf{0\%} & \textbf{92.6\%} \\
    \bottomrule
    \end{tabular}
    \caption{Ablation studies of our SphereHead framework. 
    }
    \label{tab:exper_ablation}
\end{table}
\vspace{-3mm}

\paragraph{Ablation Studies} 
\cref{tab:exper_ablation} offers a detailed assessment of each component in our framework, with gradual enhancements across different configurations. 

The transition from a tri-grid to a single-sphere representation is notable for its considerable impact, as it completely resolves mirroring-face artifacts.
The introduction of the second sphere further refines visual quality, by remedying polar and seam artifacts.
Additionally, this modification unexpectedly assists in reducing multiple-face artifacts, a fact supported by a significant decrease in fake face rate from 3.7\% to 0.9\%.
This indicates that the polar and seam regions on the single sphere are more prone to imbalanced supervision, due to their inherent discontinuity on the unfolded feature plane. Subsequently, the additional sphere compensates these regions and ensures greater overall stability.

While the combination of the single-sphere model and our view-image consistency loss gains noteworthy performance, the incorporation of an additional sphere succeeds in completely eliminating fake faces in all the generated samples within this user study, with only minimal extra parameters and computation. Lastly, integrating parsing information further mitigate artifacts and yield synthesis results with higher quality. Note that in \cref{tab:exper_ablation}, the view-image consistency loss brings substantial enhancements across the representations including tri-grid, single-sphere and dual-sphere formats. This suggests that it can be readily integrated with other frameworks for broader applicability.

\subsection{Single-view GAN Inversion}
We compare our method with PanoHead on 3D full-head reconstruction from a single-view image. 
Both methods utilize an identical GAN inversion rendering technique.
The procedure begins by ascertaining the corresponding latent code $z$ through optimization using pixel-wise L2 loss and image-level LPIPS loss \cite{zhang2018unreasonable}. This is followed by the application of Pivotal Tuning Inversion (PTI) \cite{roich2022pivotal} to alternate generator parameters, keeping the optimized latent code $z$ intact. 
As depicted in \cref{fig:teaser}, although PanoHead constructs the coarse geometry of the posterior head region, it suffers from noticeable semantic mismatch artifacts. In contrast, our method achieves an artifact-free reconstruction of both the 3D head geometry and its corresponding renderings. 

Additionally, we invited 39 participants to respond to a questionnaire that included 20 comparative scenarios. Each scenario presented the target image alongside results produced by both the PanoHead and our proposed model. Participants were tasked with determining which model generated the superior full-head visual quality and rationality. 
The study's results revealed that, barring one scenario, more respondents favored the outcomes generated by our model in all instances. Averaging across all scenarios, 82.1\% of choices favored our model's results, with only 17.9\% supporting the results from PanoHead. These findings reinforce the comparative advantage of our model over PanoHead in the realm of monocular 3D full-head reconstruction.

\section{Discussion}
\label{sec:discus}

\paragraph{Discussion on Broader Usage}
Our SphereHead demonstrates excellent feature learning capabilities for the human head, which has a spherical-like geometry. This suggests that the proposed tri-plane in spherical coordinate system is potentially adaptable to more general subjects, whose appearance exhibits strong asymmetry with respect to different directions.
Besides, the newly proposed view-image consistency loss compels the discriminator to ensure the alignment of images with their corresponding viewpoints, which is critical in 3D-aware synthesis. This paper demonstrates this emphasis particularly beneficial in the field of 3D full head synthesis, where imbalanced supervision is prominent. We anticipate its applicability to other 3D-aware GANs and broader 3D contexts, given that imbalanced supervision is a pervasive challenge across various 3D domains.

\paragraph{Limitations and Future Work}
While SphereHead successfully tackles the mirroring-face and multiple-face artifacts for 3D full head generation, it still shares some similar issues as previous 3D-aware generators, including flickering texture and lack of finer high-frequency visual and geometric details. 
We leave it as future work such as switching the backbone of the generator to StyleGAN3. 
In addition, SphereHead still suffers from data bias, especially for the back-view head. With our open-source toolbox, we're looking forward to a more diverse full-head dataset to facilitate the realm.



\paragraph{Ethical considerations}
SphereHead might be misused to manipulate the images of real people, which could pose a potential societal threat. 
We do not condone or endorse the use of our work with the malicious intent of spreading misinformation, nor do we support violating the rights of others in any form or manner.

\section{Conclusion}
\label{sec:conclu}

We propose SphereHead, a novel 3D-aware GAN framework for high-quality, view-consistent full head images generation. 
Our approach leverages a unique tri-plane representation rooted in spherical coordinates, complemented by a view-image consistency loss. 
Collectively, these innovations enable SphereHead to significantly mitigate two prevalent types of semantic mismatch artifacts, specifically the mirroring-face and multiple-face artifacts, that are commonly observed in contemporary methods.
Additionally, our method showcases superior qualitative and quantitative results against existing state-of-the-art 3D-aware GANs. 
We have also curated a comprehensive dataset, the WildHead Dataset, whose images are captured in the wild from various viewing angles spanning 360 degrees. 
We also provided an accompanying open-source data processing toolkit to support further research. We believe our SphereHead will prove beneficial to the research community in the realm of 3D full head synthesis.

\section*{Acknowledgements}
\sloppy
The work was supported in part by the Basic Research Project No. HZQB-KCZYZ-2021067 of Hetao Shenzhen-HK S\&T Cooperation Zone, Guangdong Provincial Outstanding Youth Fund (No. 2023B1515020055), the National Key R\&D Program of China with grant No. 2018YFB1800800, by Shenzhen Outstanding Talents Training Fund 202002, by Guangdong Research Projects No. 2017ZT07X152 and No. 2019CX01X104, by Key Area R\&D Program of Guangdong Province (Grant No. 2018B030338001), by the Guangdong Provincial Key Laboratory of Future Networks of Intelligence (Grant No. 2022B1212010001), and by Shenzhen Key Laboratory of Big Data and Artificial Intelligence (Grant No. ZDSYS201707251409055). It is also partly supported by NSFC-61931024, NSFC-62172348, and Shenzhen Science and Technology Program No. JCYJ20220530143604010.


\clearpage

\appendix


\begin{center}
    \Large\bfseries Supplementary Material
\end{center}

\setcounter{footnote}{0}

The supplementary material is organized as follows. 
Section \ref{supp:rep} contains the mathematical specifications of the proposed dual spherical tri-plane representation, accompanied by an analysis of this representation. 
Section \ref{supp:data} details the datasets we've compiled and the associated data pre-processing methods utilized. 
Section \ref{supp:qual} presents a more extensive range of visualizations including random sampling, single-view image rendering, and latent space interpolation. 
Section \ref{supp:video} appends a video to this supplementary material for enhanced comprehension and visualization.

\section{Dual Spherical Tri-plane Representation} 
\label{supp:rep}
\begin{figure}[h]
    \centering
    \includegraphics[width=\linewidth]{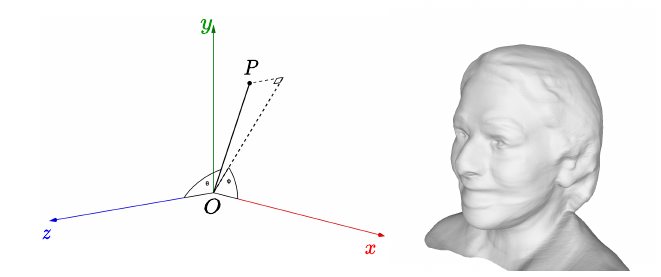}
    \caption{(a) The Cartesian coordination system $\mathcal{C}$ and the spherical coordination system $\mathcal{S}$. (b) The corresponding head pose.}
    \label{fig:supp-coord}
\end{figure}

In this paper, we render synthesis images in Cartesian coordination system $\mathcal{C}$, where a point is represented as $(x, y, z)$. 
Consequently, both the camera representation and the volume rendering module are configured and computed within $\mathcal{C}$, which aligns with EG3D \cite{chan2022efficient} and PanoHead \cite{an2023panohead}. 
The main difference is that we query a point in the 3D space via a spherical coordination system $\mathcal{S}$, retrieving its features $F_{\theta r}, F_{\phi r}, F_{\theta \phi}$ by projecting its coordinate $(r, \theta, \phi)$ onto the three feature planes, $P_{\theta r}$, $P_{\phi r}$, and $P_{\theta \phi}$ respectively.
As depicted in \cref{fig:supp-coord}, we align the front, upward, and leftward directions of the 3D head model with the positive directions of the z, y, and x-axes of $\mathcal{C}$, respectively. In the spherical system $\mathcal{S}$, the polar axis and the initial meridian plane are correspondingly aligned with the z-axis and the xoz plane of $\mathcal{C}$. This alignment ensures a coherent interaction between the two coordinate systems, facilitating the corresponding feature queries.

\subsection{Feature Query}
\begin{figure}[h]
    \centering
    \includegraphics[width=0.8\linewidth]{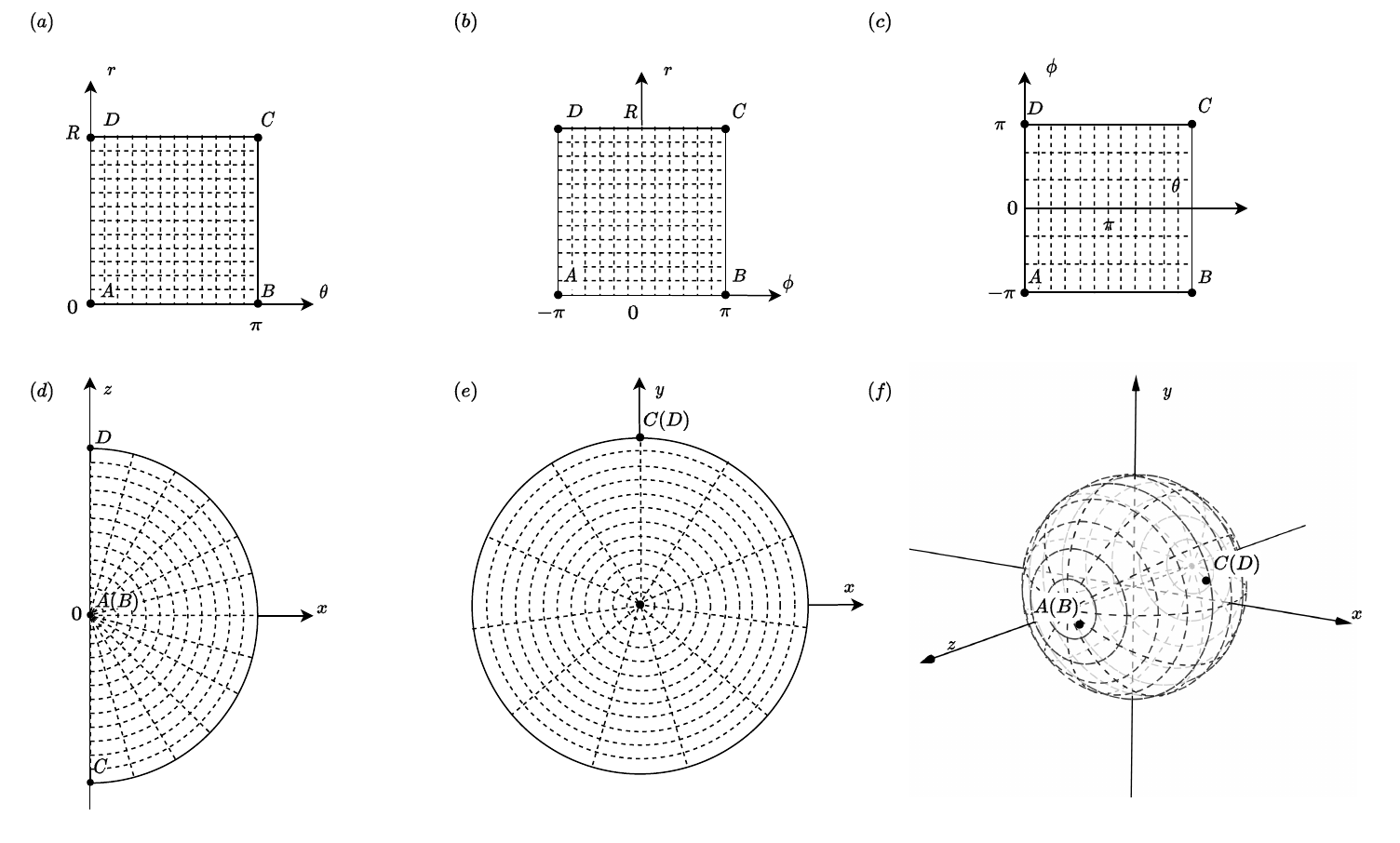}
    \caption{(a-c) The stretched feature maps $P_{\theta r}$, $P_{\phi r}$, and $P_{\theta \phi}$. (d-f) The represented geometries of $P_{\theta r}$, $P_{\phi r}$, and $P_{\theta \phi}$ in 3D space.}
    \label{fig:supp-planes}
\end{figure}

Now, we delve into the method we employ to derive the features of a given query point. Let's consider a point defined in Cartesian coordinates $(x, y, z)$ within $\mathcal{C}$. Its corresponding spherical coordinates $(r, \theta, \phi)$ within $\mathcal{S}$ can be calculated using the following equations:

\begin{equation}
    \begin{aligned}
    r &= \sqrt{x^2 + y^2 + z^2} \in \left [0, R \right ],  \\
    \theta &= \arccos\left(\frac{z}{\sqrt{x^2+y^2+z^2}}\right)  \in \left [0, \pi \right ], \\
    \phi &= \text{arctan2} \left(\frac{y}{x}\right)  \in \left [-\pi, \pi \right ],    
    \end{aligned}
\end{equation}
where $R$ constitutes the radius of the represented spherical region.
We represent the square feature map, output by the generator $G$, denoted by the square ABCD. 
During experimentation, we set the resolution of the feature map to $256\times256$. 
To cater to the different ranges of $r$, $\theta$, and $\phi$, we stretch the square feature map into three rectangles correspondingly, illustrated in \cref{fig:supp-planes} (a-c).
Lastly, to ensure their intended geometric interpretations, further transformations are applied, the results of which are depicted in \cref{fig:supp-planes} (d-f).

However, these topologically distorting and non-area-preserving transformations introduce two issues: the uneven distribution of features and the numerical discontinuity of $P_{\phi r}$ and $P_{\theta \phi}$ at $\phi=-\pi$ and $\phi=\pi$, resulting in the polar and seam artifacts respectively.
To solve them, we propose a scheme where two orthogonal spherical tri-planes interweave. 
More specifically, these planes query features via coordinates of specific spheres whose seams are orthogonal to each other. The seams of the two spheres are illustrated in Fig. 4 (e) in the main paper.
The corresponding coordinates of these two spheres correlating to their original coordinates $(r_0, \theta_0, \phi_0)$ are respectively defined as:

\begin{equation}
\left\{
    \begin{aligned}
        r_A &= r_0, \\
        \theta_A &= \arccos(\sin{\phi_0} \cdot \sin{\theta_0}), \\
        \phi_A &= \text{arctan2} (\cos{\phi_0} \cdot \tan{\theta_0}).
    \end{aligned}
\right.
\end{equation}

\begin{equation}
\left\{
    \begin{aligned}
        r_B &= r_0, \\
        \theta_B &= \arccos(-\cos{\phi_0} \cdot \sin{\theta_0}), \\
        \phi_B &= \text{arctan2} (\sin{\phi_0} \cdot \tan{\theta_0}).
    \end{aligned}
\right.
\end{equation}

To achieve a seamless blend between the spheres, we design a 2D weight map defined by the following equation:
\begin{equation}
w(\theta,\phi)=\frac{1+\cos (2\theta-\pi)}{2} \cdot \frac{1+\cos (\phi)}{2}.
\label{equ:weight}
\end{equation}
For a clearer illustration, we provide the accompanying 3D visualization code \texttt{vis\_weights.py} as a supplement to the visualizations shown in Fig. 4 (f-h).
We then use the equation below to compute the fused feature $f_F$:
\begin{equation}
f_F = \frac{w(\theta_A, \phi_A)f_A + w(\theta_B, \phi_B)f_B}{w(\theta_A, \phi_A) + w(\theta_B, \phi_B)+\epsilon},
\end{equation}
where $\epsilon$ is a very small positive number introduced to deter the denominator from equaling zero.

\subsection{Analysis of Representation}
The main advantage of the spherical tri-plane representation design lies in its fundamental ability to eradicate mirroring artifacts. Crucially, in rendering procedures, the object surface modeling plays a vital role. Our uniquely spherical conception nicely conforms to the geometric characteristics of the human head, leading to an effective aggregation of surface features on the spherical feature map $P_{\theta \phi}$. This becomes particularly indispensable during the training of a 3D full-head GAN. As articulated in Sec. 4.1 of the main paper, an issue arises from the imbalance of supervision; the features originating from regions with strong supervision tend to monopolize the shared feature plane. They subsequently permeate into regions with weaker supervision in the tri-plane and tri-grid representations. In stark contrast, our spherical feature map $P_{\theta \phi}$ rigorously differentiates features deriving from different head regions each exhibiting varied degrees of supervision. This prevents feature permeation at its root, thereby effectively eliminating mirroring artifacts.

Strictly speaking, as an efficient representation, there is still feature sharing across different locations in our design. 
Nevertheless, given the distinctive meanings of these locations, the network can easily distinguish them. 
For instance, all points along the ray $(r, \theta_0, \phi_0)$, where $r$ belongs to the range $[0, R]$, share the same $F_{\theta \phi}$. However, $F_{\theta \phi}$ only needs to concentrate on representing the features of the corresponding head surface region rather than empty areas, because the network can readily determine if a point lies on the head surface, guided by cues from $F_{\theta r}$ and $F_{\phi r}$.
Therefore, the excellent fit of $P_{\theta \phi}$ to the shape of the head surface results in a concentration of most head surface features on this plane. Thus, even though points with identical $\theta$ and $\phi$ values share features on $F_{\theta r}$ and $F_{\phi r}$, it does not cause obvious artifacts. This is because these features convey only secondary information, and their effect is hence minimal.

\section{Datasets Details} 
\label{supp:data}

\subsection{WildHead Dataset}

To leverage the abundant high quality image online, we implement a python script to scrap head images captured in the wild from various viewing angles online. 
We adopt the open-source \texttt{icrawler} library and search for keywords on the Internet. 
For female images, our searching keywords include \textit{Bob cut, Pixie cut, Beach Waves, Bangs}, etc.; for male images, we specify keywords including \textit{Buzz cut, Bowl cut, Undercut, Bald}, etc. 

We employ our data processing toolbox to filter out images that are occluded, low-resolution, or blurred. Subsequently, we crop and align each image to ensure consistency and quality.

Considering privacy concerns, we have processed the front-facing images in our dataset using a publicly accessible face swapping api\footnote{https://huggingface.co/spaces/felixrosberg/face-swap} to anonymize the facial identities present in the original data. To maintain the quality of the dataset, we manually removed any images that appeared unnatural or contained artifacts post-processing. Additionally, we provide a mechanism for image removal requests. If you believe any image in the dataset infringes on your privacy, please contact us for immediate removal\footnote{Email us at \href{mailto:heyuanli@link.cuhk.edu.cn}{heyuanli@link.cuhk.edu.cn}}.

After the above processing, the final dataset consists of 60k images.

\subsection{Data Processing}

For each obtained image, we begin by filtering it based on its size. Images whose width or height is smaller than 512 pixels are discarded.

Next, we apply two blur detection methods to the image. The first method, proposed in \cite{su2011blurred}, calculates a blur degree, where 0 indicates clarity and 1 indicates blur. In our implementation, we use a threshold of 0.75. This method detects blurred regions by analyzing the singular value information of each pixel in the image. The second method, proposed in \cite{pech2000diatom}, applies the Laplacian operator to the image and computes the variance of the results. Higher variance values indicate clear images, while lower values suggest blurriness. We set a threshold of 50 for this method.

Following the blur detection step, we utilize a YOLO detector \cite{zhang2021soft} to identify heads within the image. If no head box is found, the image is discarded.

Once we have the head box, we apply a face landmark detector \cite{bulat2017far} to the head region. If the landmark detection result is not None, the corresponding image is labeled as a front view image; otherwise, it is marked as a back view image.

For the front view images, we align them using the landmarks, following the alignment process proposed in PanoHead~\cite{an2023panohead}. During this alignment, we record the aligned quad box and calculate the transformation (scale, rotation, and translation) from the head detection box to the aligned quad box. Additionally, we employ 3DDFA\_V2 \cite{deng2019accurate} to estimate the camera's world-to-camera rotation matrix for this image.

After processing all the front view images, we calculate the mean transformation from the head box to the aligned box. For the back view images, we apply this transformation to each image's head box to obtain the corresponding aligned box. We then crop the image using the calculated aligned box. Subsequently, WHENet \cite{zhou2020whenet} is utilized to estimate the pitch, yaw, and roll angles of the head. 
To guarantee accuracy in head pose representation, we perform a manual review and make necessary adjustments to the pose of each image.
These angles then facilitate the derivation of the camera's world-to-camera rotation matrix.

Regarding the camera parameters, we adopt the method described in PanoHead. Since each image is aligned by cropping with the align quad box, ensuring that the subjects are positioned at the center, we assume that human heads are located at the center of the world coordinates. The cameras are situated on the surface of a sphere with a radius of 2.7m, centered at the origin. All cameras share the same intrinsic matrix $C$, which is defined as follows:

\begin{equation}
    C=\begin{bmatrix}
4.2647 & 0 & 0.5 \\
0 & 4.2647 & 0.5 \\
0 & 0 & 1
\end{bmatrix}
\end{equation}

Since all cameras face inwards toward the origin point of the world, the extrinsic matrix $P$ can be expressed as:

\begin{equation}
    P=\begin{bmatrix}
I & t \\
0 & 1
\end{bmatrix}\begin{bmatrix}
R & 0 \\
0 & 1
\end{bmatrix}, \quad t=\begin{bmatrix}
0 \\
0 \\
-2.7
\end{bmatrix}
\end{equation}

We then employ DeepLabV3 \cite{chen2017rethinking} to obtain the background and foreground masks for each image. Regarding parsing, as current parsing models yield inferior results for back view images, we employ three parsing models (one is from \cite{yu2021bisenet} and the others are from \cite{lin2021roi} with different backbone) to vote for the final semantic label of each pixel. Consequently, we generate parsing results for each image.

Subsequently, we calculate the ratio of the foreground area to the entire mask and the ratio of valid semantic pixels to all pixels. Both ratios have a threshold of 0.3. Images with ratios smaller than 0.3 are discarded.

In order to rectify the imbalance in camera positions following dataset preparation, we allocate a duplication count $N_{\text{dup}}$ to each viewpoint ($\theta$, $\phi$). $N_{\text{dup}}$ is determined based on the quantity of images available at the azimuth angle $\theta$.

\begin{equation}
    N_{dup}=\begin{cases}1&(N_\theta\geq N_{thresh})\\\texttt{ceil}(N_{thresh}/N_\theta)&(N_\theta< N_{thresh})\end{cases},
\end{equation}
where $N_{\text{thresh}}$ is a constant set to 2000. By this duplication strategy, we balance the distribution of view angles in the training dataset.

\section{Additional Qualitative Results and Analysis} 
\label{supp:qual}

\subsection{Random Sampling} 

\begin{figure*}
    \centering
    \includegraphics[width=\linewidth]{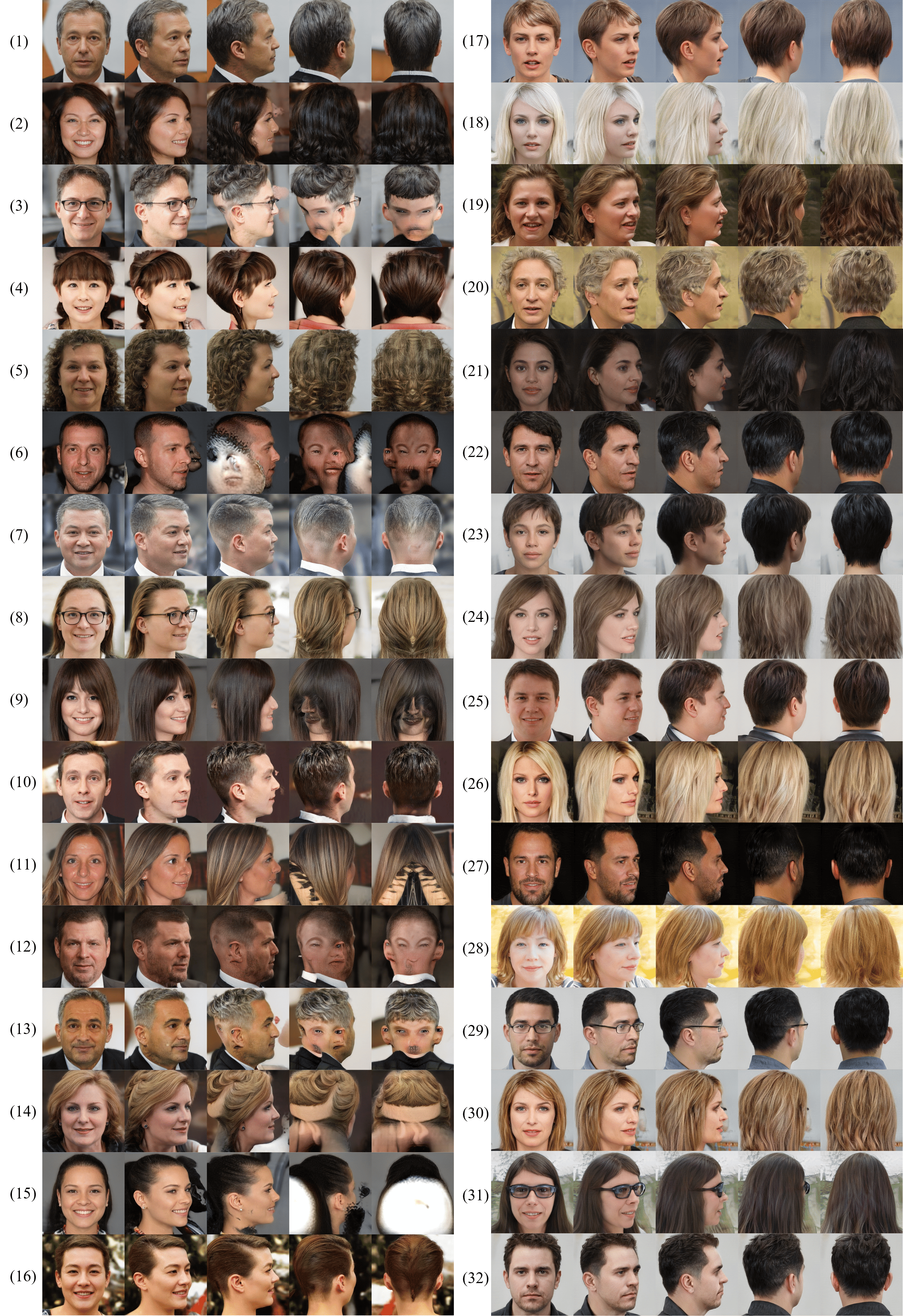}
    \caption{Comparison of random sampling. Left column (1-16): results of the officially released PanoHead model \cite{an2023panohead}. Right column (17-32): results of our SphereHead. }
    \label{fig:supp-random1}
\end{figure*}

We conduct a comparative analysis of 3D full-head synthesis capabilities between our SphereHead model and PanoHead \cite{an2023panohead}, which is the strongest baseline. \cref{fig:supp-random1} presents this comparison, utilizing the officially released PanoHead model that has been trained on the FFHQ-F dataset. Given that the in-house data included in the FFHQ-F dataset is not publicly accessible, we have developed an image collection and trained our model on it, as elaborated in 
Sec. 5 in the main paper.

\begin{figure*}
    \centering
    \includegraphics[width=\linewidth]{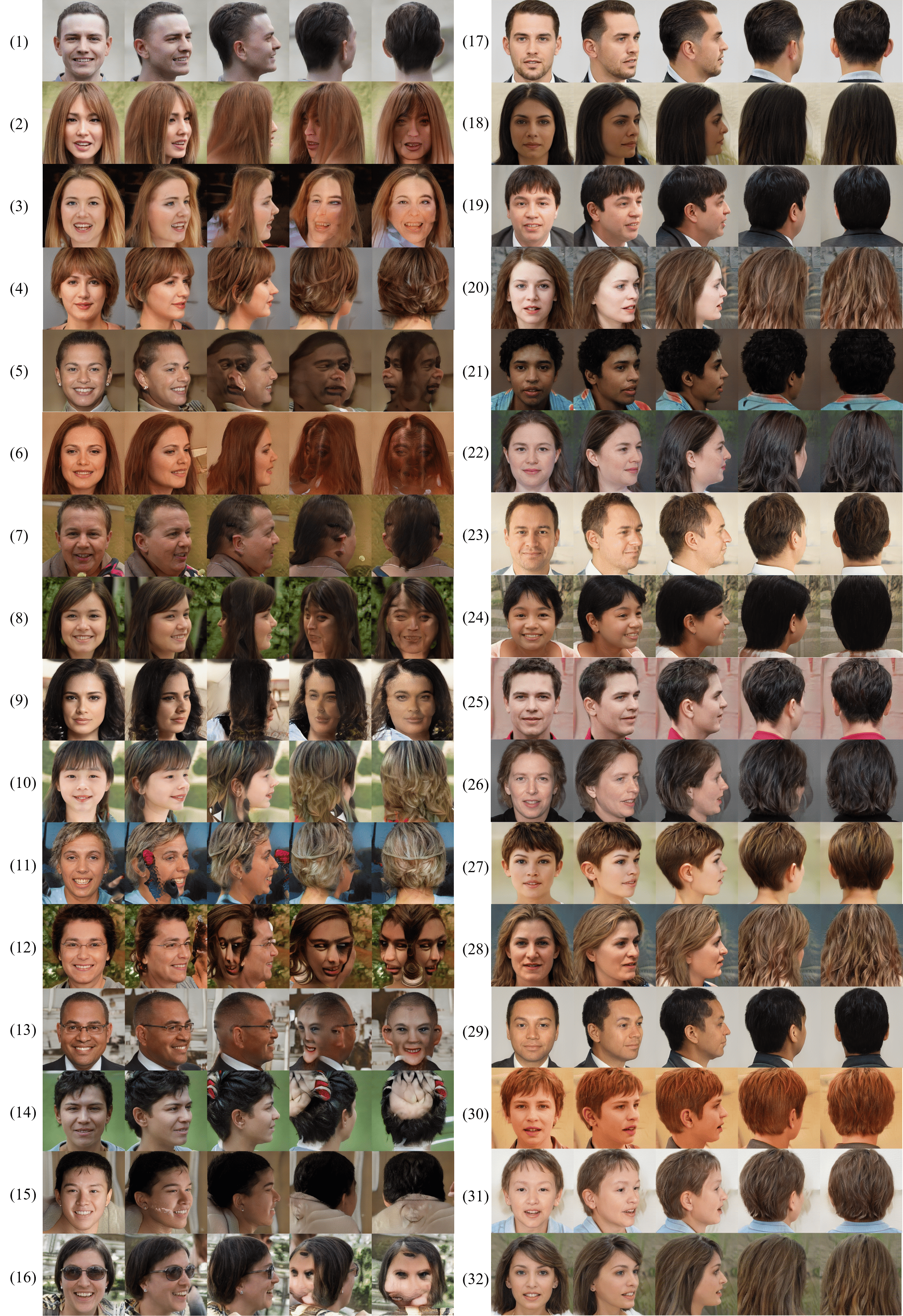}
    \caption{Comparison of random sampling. Left column (1-16): results of PanoHead \cite{an2023panohead}. Right column (17-32): results of our SphereHead. Both models are trained from stretch using our image collection.}
    \label{fig:supp-random2}
\end{figure*}

To ensure a fair comparison, we also trained a PanoHead model using their publicly accessible codebase\footnote{https://github.com/SizheAn/PanoHead} on our data. The corresponding comparison is depicted in \cref{fig:supp-random2}. Our observations indicate that, versus the officially released model, the strict left-right symmetry phenomena are mitigated in the back-view synthesis of the PanoHead model trained on our data. We attribute this reduction to the increased volume of back-view images in our collection, which showcases a rich diversity of hairstyles exhibited under a broad spectrum of head poses and photographic circumstances.

Nevertheless, the back-view outputs of the model trained on our data exhibit a degree of disarray, as if distinctive hairstyles are tangled together, which is observable in \cref{fig:supp-random2}(4, 7, 10, 11). 
This outcome is primarily attributed to the inclusion of more complex and diverse hairstyles in our dataset.
Initially, we presumed that incorporating such varied hairstyles would enhance the representational capacity of a comprehensive full-head model. However, it now appears that these complex patterns might introduce noise, particularly if the model's capacity is insufficient to accurately manage them.

As depicted in \cref{fig:supp-random1} and \cref{fig:supp-random2}, our SphereHead model effectively eliminates pervasive issues seen in PanoHead, such as strict left-right symmetry (\cref{fig:supp-random1}(2, 4, 5, 6, 7, 8, 11, 16) and \cref{fig:supp-random2}(4)) and mirroring-face artifacts (\cref{fig:supp-random1}(6, 9) and \cref{fig:supp-random2}(2, 3, 6, 8, 9)) in the back-view images. In addition, SphereHead also gets rid of multiple-face artifacts (\cref{fig:supp-random1}(3, 12, 13, 14) and \cref{fig:supp-random2}(5, 12, 16).
Moreover, additional minor artifacts such as uneven coloration, contorted hair strands, and peculiar line formations in the results from PanoHead can be traced back to subtler forms of mirroring-face and multiple-face artifacts. For instance, the patch of darker shading on the occipital region in \cref{fig:supp-random1}(7), the ocular resemblances adjacent to the ears in \cref{fig:supp-random2}(7), and the facial contour lines on the shoulders in \cref{fig:supp-random2}(15) are indicative of these less pronounced mirroring-face and multiple-face artifacts.
In contrast, by completely eliminating these artifacts, the outputs generated by our SphereHead model appear noticeably more realistic and natural.

Besides, the outcomes from our models exhibit noticeably fewer floating and other obvious artifacts when compared with those of PanoHead, such as \cref{fig:supp-random1}(6, 11, 14, 15) and \cref{fig:supp-random2}(11, 14). This improvement is presumably due to our spherical representation closely fitting the geometry of human heads, which yields beneficial effects. Firstly, it simplifies the process for the feature maps $F_{\theta r}$ and $F_{\phi r}$ to distinguish between the internal and external regions of the head within the spatial domain. 
Secondly, it enables $F_{\theta \phi}$ to more accurately grasp the features' semantic meanings on the spherical layout. 
Both factors are instrumental in the observed reduction of artifacts.

\subsection{Single-view image Inversion} 
We undertake a comparative analysis between our proposed model and the PanoHead model \cite{an2023panohead}, focusing on monocular 3D full-head reconstruction. 
This task presents a significant challenge to both models as the only available constraints are derived from the front view, leading to side-view and back-view reconstructions that manifest as rather blurred. Essentially, all side-view and back-view reconstruction constraints arise from each model's respective latent space. Therefore, this serves as an appropriate test of each model's capability to infer the head's back view and provides a means to evaluate the efficacy of the latent spaces within the two models.

In this comparison, the target images are randomly sampled from the FFHQ dataset \cite{karras2018style}. Both models employ the same inversion technique for this comparison. 
The inversion begins with optimizing the corresponding latent code $z$ via pixel-wise L2 loss and image-level LPIPS loss \cite{zhang2018unreasonable}. This is succeeded by implementing Pivotal Tuning Inversion (PTI) \cite{roich2022pivotal} to adjust the generator parameters while leaving the optimized latent code $z$ unaffected. 

\begin{figure*}
    \centering
    \includegraphics[width=0.8\linewidth]{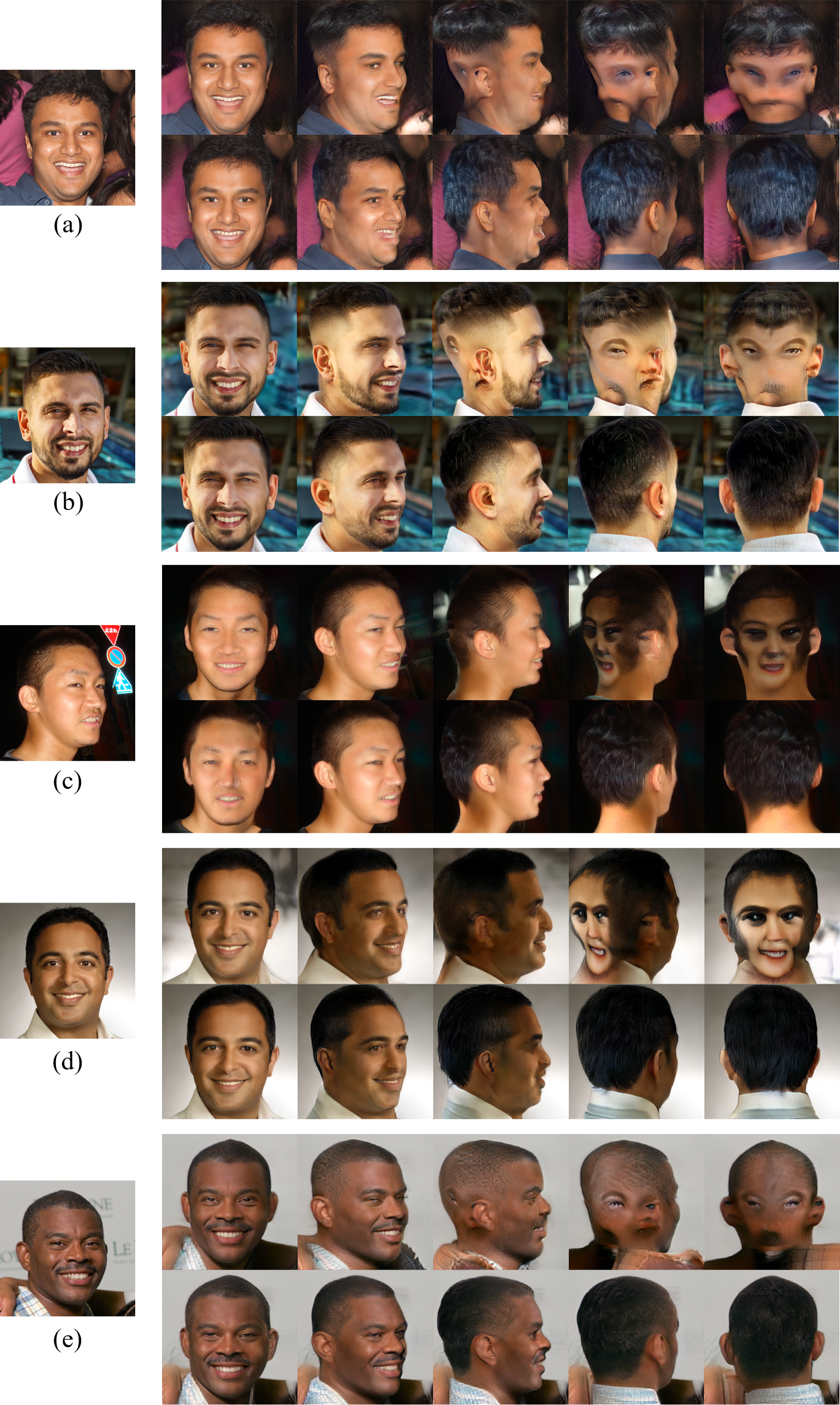}
    \caption{Comparison of single-view image inversion. 
    The monocular images on the left are target images from FFHQ dataset \cite{karras2018style}. The upper and lower rows on the right portray are the results from PanoHead \cite{an2023panohead} and our SphereHead model, respectively.}
    \label{fig:supp-inverse1}
\end{figure*}

\begin{figure*}
    \centering
    \includegraphics[width=0.8\linewidth]{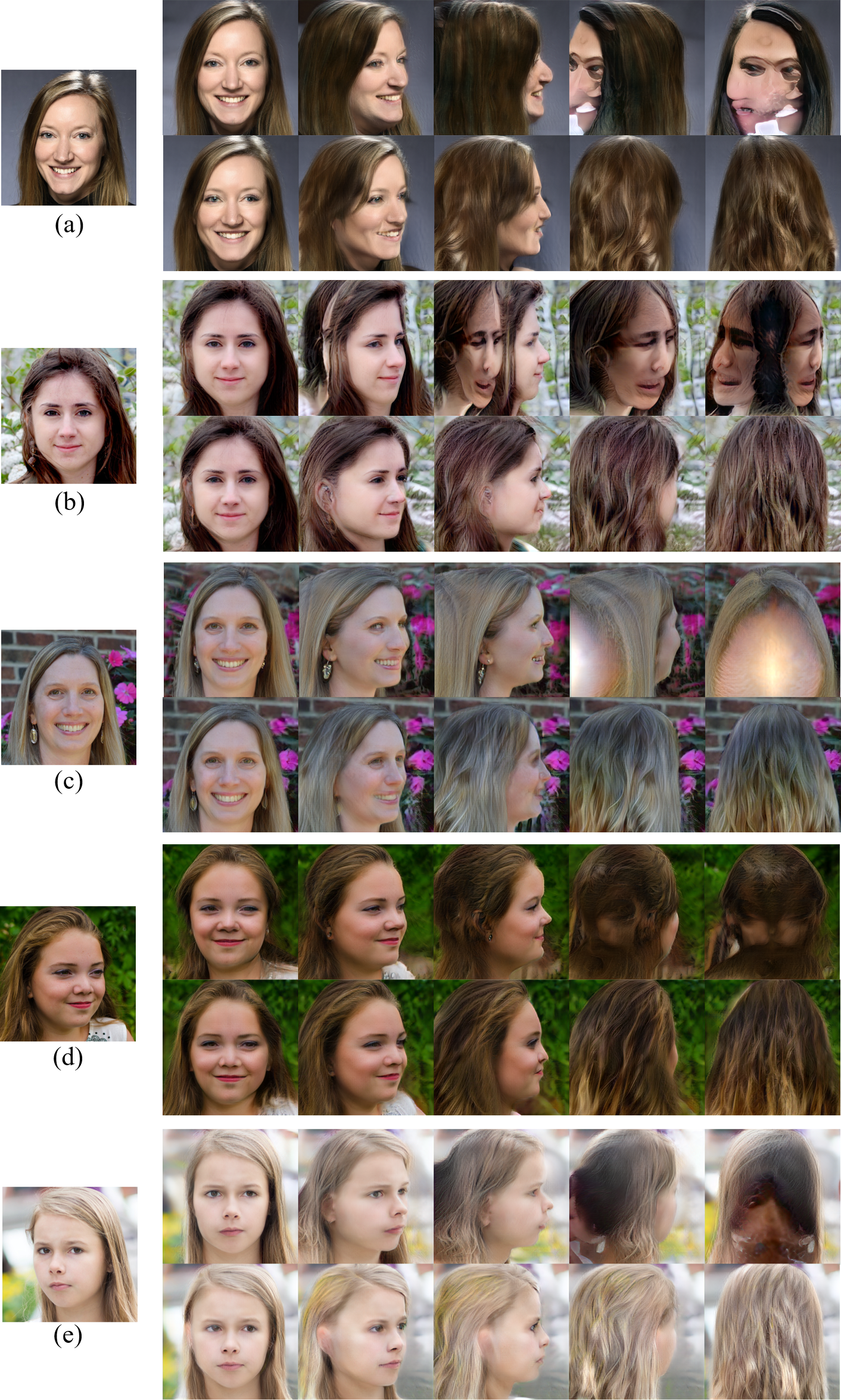}
    \caption{Comparison of single-view image inversion. 
    The monocular images on the left are target images from FFHQ dataset \cite{karras2018style}. The upper and lower rows on the right portray are the results from PanoHead \cite{an2023panohead} and our SphereHead model, respectively.}
    \label{fig:supp-inverse2}
\end{figure*}

As demonstrated in \cref{fig:supp-inverse1} and \cref{fig:supp-inverse2}, given the same target samples, both models can successfully reconstruct valid front-view images. However, PanoHead's back-view synthesis typically fails, leading to mirroring-faces and multiple-face artifacts (\cref{fig:supp-inverse1}(a,b,c,d,e) and \cref{fig:supp-inverse2}(a,b,d,e)), as well as excessive blurring and an absence of detail(\cref{fig:supp-inverse1}(e). In contrast, our proposed method consistently generates accurate and detailed back-view reconstructions.

We further substantiate our model's efficacy by conducting a comparative user study on single-view image inversion. We invited 39 participants to respond to a questionnaire that included 20 comparative scenarios. Each scenario presented the target image alongside results produced by both the PanoHead and our proposed model, akin to the format in \cref{fig:supp-inverse1}. Participants were tasked with determining which model generated the superior full-head visual quality and rationality. 
The study's results revealed that, barring one scenario, more respondents favored the outcomes generated by our model in all instances. Averaging across all scenarios, 82.1\% of choices favored our model's results, with only 17.9\% supporting the results from PanoHead. These findings reinforce the comparative advantage of our model over PanoHead in the realm of monocular 3D full-head reconstruction.

\subsection{Latent Space Interpolation} 
\begin{figure*}
    \centering
    \includegraphics[width=\linewidth]{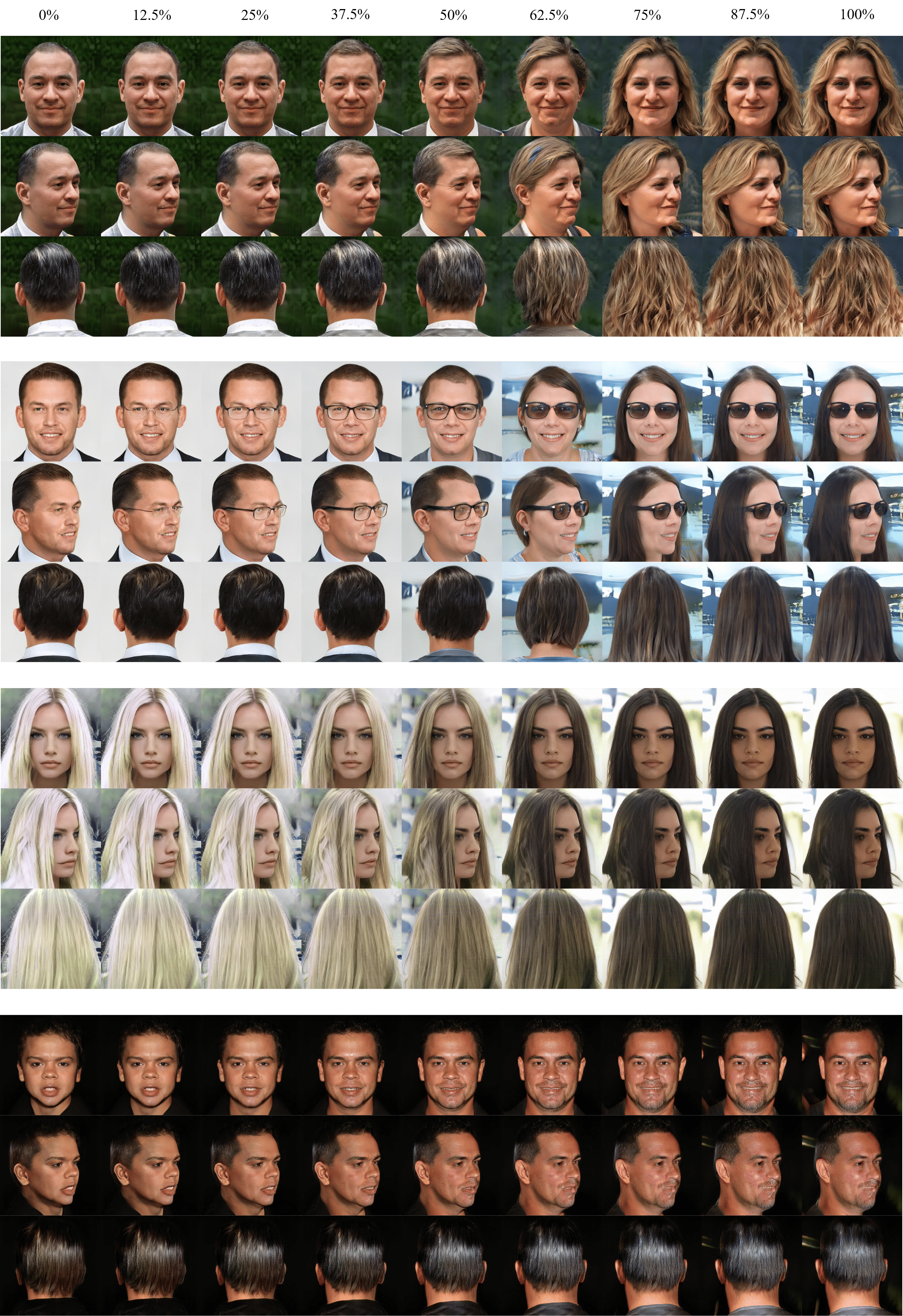}
    \caption{Visualization of latent space interpolation. 
    The extreme left and right images depict the end points of the interpolation, while the seven intermediary images represent ratio-varied interpolations.}
    \label{fig:supp-interpolation}
\end{figure*}

Presented in \cref{fig:supp-interpolation} is a depiction of how transformations in the generated 3D head visuals correlate with corresponding interpolations within the model's latent space. The process initiates with a random selection of two points within our model's latent space. These points are then amalgamated at linear ratios. The series of images, oriented from left to right, illustrate a continual, fluid semantic evolution from the first to the second identity. The seamless transitions, along with the retention of semantic coherence between interpolations, serve to exemplify the continuity and smoothness of the latent space as learned by our model.

\section{Supplementary Video} 
\label{supp:video}
As a complement to this paper, we offer an accompanying video demonstrating the 360-degree visual quality and view-consistency achieved by our model. This supplementary video serves to display additional scenarios encompassing random sampling, single-view image inversion, and latent space interpolation. We strongly recommend readers to view this material as it will enhance their understanding and provide a more concrete visualization of the concepts discussed within the manuscript.

%
%
\bibliographystyle{splncs04}
\bibliography{main}
\end{document}